# FFusionCGAN: An end-to-end fusion method for few-focus images using conditional GAN in cytopathological digital slides

Xiebo Geng[a,#], Sibo Liu[a,#], Wei Han[a], Xu Li[a], Jiabo Ma[a], Jingya Yu[a], Xiuli Liu[a], Sahoqun Zeng[a], Li Chen[b,*], Shenghua Cheng[a,*]

[a] *Britton Chance Center for Biomedical Photonics, Wuhan National Laboratory for Optoelectronics-Huazhong University of Science and Technology, Wuhan, 430074, China*
[b] *Department of Clinical Laboratory, Tongji Hospital, Huazhong University of Science and Technology, Wuhan, 430030, China*

[#] Equal contribution to this work
[*] Corresponding author: chengshen@hust.edu.cn (S. Cheng), chenliisme@126.com (L. Chen)
Email address: m201872965@hust.edu.cn (X. Geng); liam.liusibo@gmail.com (S. Liu); M201872912@hust.edu.cn (W. Han); goodlixu@163.com (X. Li); majiabo@hust.edu.cn (J. Ma); jingyayu@hust.edu.cn (J. Yu); xlliu@mail.hust.edu.cn (X. Liu); sqzeng@mail.hust.edu.cn (S. Zeng); chenliisme@126.com (L. Chen); chengshen@hust.edu.cn (S. Cheng)

**Abstract**

Multi-focus image fusion technologies compress different focus depth images into an image in which most objects are in focus. However, although existing image fusion techniques, including traditional algorithms and deep learning-based algorithms, can generate high-quality fused images, they need multiple images with different focus depths in the same field of view. This criterion may not be met in some cases where time efficiency is required or the hardware is insufficient. The problem is especially prominent in large-size whole slide images. This paper focused on the multi-focus image fusion of cytopathological digital slide images, and proposed a novel method for generating fused images from single-focus or few-focus images based on conditional generative adversarial network (GAN). Through the adversarial learning of the generator and discriminator, the method is capable of generating fused images with clear textures and large depth of field. Combined with the characteristics of cytopathological images, this paper designs a new generator architecture combining U-Net and DenseBlock, which can effectively improve the network's receptive field and comprehensively encode image features. Meanwhile, this paper develops a semantic segmentation network that identifies the blurred regions in cytopathological images. By integrating the network into the generative model, the quality of the generated fused images is effectively improved. Our method can generate fused images from only single-focus or few-focus images, thereby avoiding the problem of collecting multiple images of different focus depths with increased time and hardware costs. Furthermore, our model is designed to learn the direct mapping of input source images to fused images without the need to manually design complex activity level measurements and fusion rules as in traditional methods. The experimental results show that our method is superior to current image fusion technologies and classical image-to-image conditional GAN methods in both visual quality and objective evaluation metrics.

## 1. Introduction

In the field of cytopathological slide image analysis for cervical cancer, digital slides are the most widely used analytical processing object. However, this kind of visual analysis object rarely has all the cells in focus, which increases the challenges to cytopathologists' interpretation and further automated processing. Image fusion technologies compress multiple images with different focus depths in the same field of view into a single focused image (as shown in Fig. 1), which can effectively solve the problem where only part of the regions are in focus [1]. The purpose of image fusion is to extract the in-focus details from multiple input source images and preserve them in the final fused image [2]. Compared with single-focus images, fused images are more informative for humans or machine algorithms. Fused images help the cytopathological analysis of doctors and the development of automatic lesion cell recognition algorithms [3].

Studies in the field of image fusion have been conducted for more than 30 years, during which many image fusion methods have been proposed [3]. These image fusion schemes include multi-scale decomposition-based methods [4-11], sparse representation-based methods [12-16], spatial domain transform-based methods [17, 18], hybrid transform-based methods [19, 20], deep convolutional neural network-based methods [21, 22], and some other methods [23, 24]. Except the deep learning-based approaches, most traditional fusion frameworks basically consist of three key factors: the image transform, the activity level measurement and the fusion rule design [3]. All of the above steps require manual design. In order to obtain better fusion effects, these fusion frameworks are becoming increasingly more complex. Considering the implementation difficulty and computational burden, it is very difficult to manually construct a perfect solution that can consider all important factors [25]. Current most deep learning based-fusion methods are only used to replace a part of traditional methods rather than directly conducting end-to-end mapping. For example, the method in [21] learned a convolutional neural network (CNN) to replaces the activity level measurement in traditional methods. Furthermore, currently, both traditional image fusion methods and deep learning-based fusion methods need to capture multiple input source images of different focus depths in the same field of view in advance, which requires large time and hardware costs. The problem is especially prominent in large-size cytopathological whole slide images (WSIs). To obtain clear fused WSIs, it is necessary to adjust the focus depth and scan WSIs multiple times (9 to 21) due to the characteristics of cytopathological cells, which costs a lot.

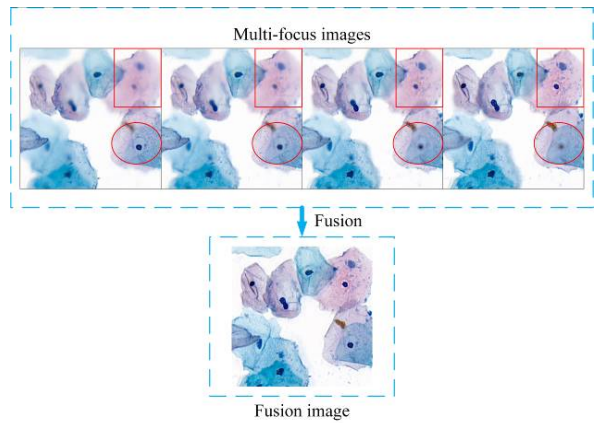

**Fig. 1**. Multi-focus image fusion. The multi-focus images in the figure are digital cytopathological images of cervical cells that were taken at different focus depths. The red rectangular area gradually becomes clear from left to right, and the red elliptical area gradually becomes blurred from left to right. The in-focus regions vary at different focus depths. The fused image that combines the multi-focus images can basically keep all the objects in focus.

To address the above challenges, in this paper, we proposed a novel end-to-end method to generate fused images from single-focus or few-focus images (≤3) based on conditional GAN. Through the adversarial learning of the generator and discriminator, our method can generate fused images with clear textures and large depth of field from single-focus or few-focus images. Due to the randomness of the focal and non-focal cell distribution in cytopathological images, this paper designs a generator architecture combining U-Net and DenseBlock, which can effectively expand the network's receptive field and comprehensively encode image features. Meanwhile, this paper develops a semantic segmentation network that identifies the blurred regions in cytopathological images. By integrating the output mask of the network into the above generative model, the generator can perform high-quality fusion of images for the locally blurred regions. Through a series of experiments, it is proved that the proposed method is superior to current image fusion methods and the classical image-to-image conditional GAN models. The method in this paper learns an end-to-end mapping of the input source images and fused image, thereby avoiding the design of complex activity level measurement and fusion rules. In addition, we can generate fused images from single-focus or few-focus images without having to spend significant time and hardware costs to capture multiple focus images as in traditional methods.

The main contributions of the work are the following aspects. First, we proposed an end-to-end conditional GAN model to generate fused images with clear textures and large depth of field from single-focus or few-focus images, thereby avoiding the time and hardware costs of traditional fusion methods when capturing multi-focus images. The method has important practical significance for large-size cytopathological WSIs. Second, we developed a segmentation network that identifies the blurred regions in cytopathological images, and uses the output mask of the network as an additional condition of the generator. Experiments show that the scheme can improve the quality of the generated fused images. Third, by establishing the direct mapping between the input source images and output fused images, the model avoids the manual design of complex activity level measurement and fusion rules. The content of the remainder of this paper is arranged as follows. Section 2 describes the work that is related to multi-focus image fusion in detail. Section 3 describes the proposed method of generating a fused image from single-focus or few-focus images based on conditional GAN. The experimental results of the proposed method are given in Section 4, including the experimental data, the training details, the quantification of results, and the comparison with the traditional fusion algorithms, deep learning-based fusion algorithms and the classic image-to-image conditional GAN models. The Section 5 will discuss and summarize the work of this paper.

## 2. Related work

In this section, we will introduce some background materials and related work in detail, including traditional multi-focus image fusion methods and deep learning-based multi-focus image fusion methods, and finally we will explain the difference between the proposed method and these related methods.

*2.1. Traditional multi-focus image fusion methods*

The research literature on traditional multi-focus image fusion is very rich. According to [3], the current traditional image fusion methods can be classified into the following categories: multi-scale decomposition-based methods, sparse representation-based methods, spatial domain-based methods and hybrid transformation-based methods. We will briefly describe the ideas of these image fusion methods.

Typical multi-scale decomposition-based fusion methods compress multiple input images into a fused image through a three-stage framework: multi-scale decomposition, activity level measurement and multi-scale reconstruction [25]. Common multi-focus image fusion methods based on multi-scale decomposition include the wavelet transform method [5] and the double-tree complex wavelet transform method [6], the Laplacian pyramid method [7], multi-scale geometric transformation methods [9, 10] and so on. The main idea of sparse representation-based fusion methods is that the activity level of the input source images can be measured in a sparse domain. In the spatial domain-based fusion methods, most are based on image division or image segmentation [17, 18]. However, these image fusion methods have concurrent advantages and disadvantages. For example, fusion methods based on multi-scale decomposition can extract the spatial structures of different scales in the input images but cannot represent the low-frequency components sparsely. Fusion methods based on sparse representation can give a meaningful representation of the input source images using dictionary learning. But it has difficulties when reconstructing the small-scale details of the input source images due to the limited number of atoms in the dictionary [3]. To combine the advantages of these methods, some complicated hybrid image fusion models have emerged. For example, Yu et al. proposed a hybrid method for multi-focus image fusion using wavelets and classifiers [19]. In short, traditional fusion methods are designed to identify the in-focus regions of the input source images and preserve them in the output fused image. However, they cannot produce extra information other than the input images.

*2.2. Multi-focus image fusion methods based on deep learning*

In recent years, with the widespread use of deep learning, increasingly more researchers have applied deep convolutional neural networks to the field of image fusion. Liu et al. adopted a deep CNN method to replace the activity level measurement of traditional methods, which overcomes the difficulties that are faced by the traditional fusion methods [21]. In [22], a pixel-based CNN is proposed that can identify the in-focus and out-of-focus pixels in the input source images. Both of these methods essentially use CNNs to replace the activity level measurement module in the traditional image fusion algorithms. The method in [23] introduced the convolution sparse representation to improve limited detail preservation ability of traditional sparse representation-based fusion methods. This approach essentially replaces the traditional image decomposition module with CNNs. In short, the existing multi-focus image fusion methods based on deep learning mostly replace a part of traditional fusion methods, similar to in [21], [22] and [23], rather than directly learning an end-to-end mapping of the input images and the fused image. Moreover, these methods are just image fusion and cannot obtain more information than what is in the input images.

There were a few methods for generating fused images using GAN. FusionGAN used GAN to fuse infrared images and visible images [26]. The infrared images and visible images are directly stacked and input into the generator and the generated images have both the overall contour of the infrared images and the details of the visible images. However, unlike the task in this paper, FusionGAN fuse infrared images and visible images instead of multi-focus images. It cannot be guaranteed that FusionGAN has good performance in the multi-focus fusion task.

*2.3. Differences between our method and other GAN-based methods*

The few-focus image fusion method proposed in this paper is based on conditional GAN. Our method and FusionGAN differ in the following two aspects. First, in terms of the problem to solve, FusionGAN is not appropriate for multi-focus image fusion, but is used to fuse visible images and infrared images. Its purpose is to fuse the details of visible images and the global information of infrared images. The purpose of our method is to incorporate the in-focus regions in the input images into the fused image as much as possible and to recover all out-of-focus regions as in-focus regions. Second, in terms of the network architecture, we designed a U-Net generator with DenseBlocks while FusionGAN's generator adopts the ordinary CNN. We designed this generator because we think that image fusion not only needs to take into account the detailed texture information of input images but also the large scale content information. In addition, PatchGAN discriminator is adopted in our method, which is different from FusionGAN's plain discriminator. PatchGAN outputs an $N \times N$ feature map. The value at each position of the feature map corresponds to the probability that each tile in the input images is true. PatchGAN's discriminator can produce sharper reconstructed images than the plain discriminator [27].

The method in this paper is an image-to-image generation method based on conditional GAN. It is similar to the typical GAN-based image-to-image model, Pixel2Pixel [27]. The generators both adopt the U-Net structure, and the discriminators adopt PatchGAN. The difference is that we replace the naive convolution structure in the U-Net with DenseBlocks [28], which can better transfer and encode features.

Finally, we designed a blurred region recognition model labeled as BM in our method. Unlike FusionGAN and Pixel2Pixel schemes, the input of our generator is the original few-focus images stacked with their corresponding blur masks instead of only the original images. We believe that the blur masks can help to reconstruct the locally blurred regions in the original images. Subsequent experiments will prove that the blur recognition module that we added is indeed valid.

In this paper, we use FusionGAN and Pixel2Pixel to train few-focus image fusion networks. That is, the networks' inputs are changed to few-focus images. Then, the results of FusionGAN and Pixel2Pixel are quantitatively and qualitatively compared with those of our method. The experimental results show that our method achieved better results.

## 3. Method

Here, we first proposed a conditional GAN that can generate a fused image from single-focus or few-focus cervical cell images. Then, the model structure is discussed, including the generator, the discriminator, a pre-trained blurred region recognition model, and the losses of the conditional GAN. Finally, we will describe the training strategy of the entire model.

*3.1. Problem formulation*

Our network architecture is shown in Fig. 2. The Blur_Model (BM) is a pre-trained blur region recognition model, that learns a mapping from image $I_m$ to blur mask $I_b$ (binary mask, where white regions are the blurred areas and black regions are the non-blurred areas). The generator $G_\theta$: $I_c\{I_m, I_b\} \rightarrow I_f$ learns to generate a fused image from the stacked images $I_c$ of $I_m$ and $I_b$. The task of the discriminator $D_\theta$ is to distinguish the real fused image $I_r$ and the generated fake fused image $I_f$ as much as possible. For a single-focus image to generate a fused image, the input of the generator is a stack of the single-focus image and its corresponding blur mask $I_b$.

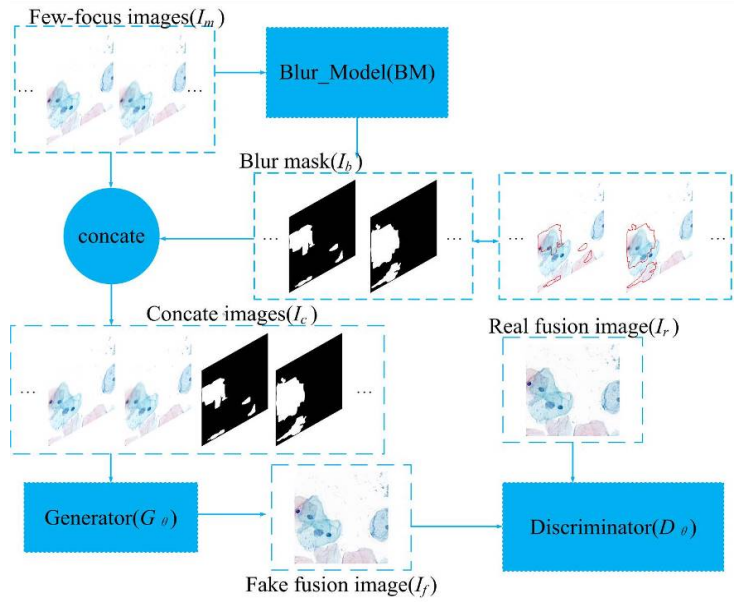

**Fig. 2**. The architecture for generating fused images from single-focus or few-focus images.

*3.2. Network architecture*

Our network architecture consists of three parts: the generator $G_\theta$, the discriminator $D_\theta$, and the semantic segmentation network BM for identifying blurred regions.

*3.2.1. Generator $G_\theta$*

We designed a new generator $G_\theta$, which is a U-Net with DenseBlocks, as shown in Fig. 3. $G_\theta$ is an improved U-Net network. U-Net is an encoder-decoder type of network structure. In the encoding phase, the global information of the input images is obtained using convolution and pooling. In the decoding phase, the global information and the local information are merged using the skip-connect stacking feature maps of different scales. Thus, the generated fused image includes the low-frequency global information and high-frequency texture information. Further, we modified the plain convolutional layers in U-Net into a DenseBlocks module, which enhances the transfer of the features and makes more efficient use of features compared to common convolutional layers.

The input of $G_\theta$ is a multi-channel image with a width of 512 and a height of 512. First, it will pass through a convolution layer with PRelu outputting a 64-channel feature map. Then, there will be three DenseBlocks and three average pooling layers. The DenseBlocks in the encoding stage (the left branch in Fig. 3) consists of a common convolution layer and two DenseBlock blocks. Unlike the DenseBlock in [28], which has a 5-layer convolution, the DenseBlock in our network only use a 3-layer convolution, thereby reducing the computational overhead. When reaching the bottom of U-Net, it will go through the bottom layer, which is a 3-layer convolution with PRelu. During the decoding stage, the network goes through three bilinear upsampling layers, and each layer is followed by a DenseBlocks. Different from the encoding stage, the input of the DenseBlocks in decoding stage is concatenated by the output of the bilinear upsampling layer and the corresponding DenseBlocks output of the encoding stage. At the top of the network, the convolutional layer and the active tanh layer are used to reduce the number of channels of the feature map and normalize it to between 0 and 1, thereby obtaining a 3-channel output.

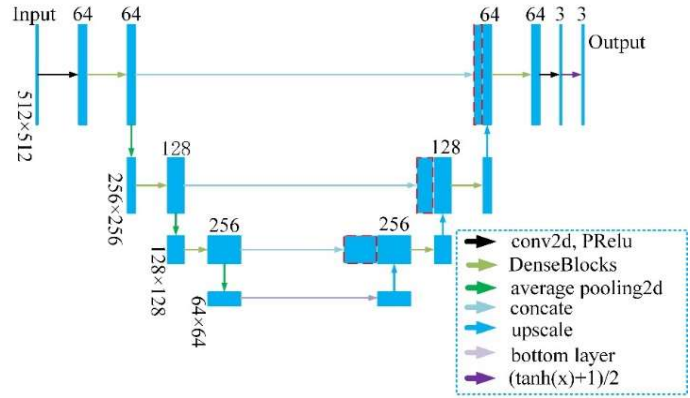

**Fig. 3**. The generator of our method, which is an improved U-Net network. 'conv2d' is the convolution with a kernel size of 9, a stride of 1 and a padding of 4. DenseBlocks is composed of multiple DenseBlock blocks. 'average pooling2d' is an average pooling layer with a stride of 2. 'upscale' is a bilinear upsampling layer with a scale of 2. 'bottom layer' is a 3-layer convolution with PRelu.

*3.2.2. Discriminator $D_\theta$*

Our PatchGAN discriminator $D_\theta$ uses a structure similar to the VGG. $D_\theta$ consists of 8 convolution layers with LeakyReLU, a maximum pooling layer with a window size of 4 and a stride size of 4, and a sigmoid activation layer. $D_\theta$ outputs an $N \times N$ map with values between 0 and 1. The receptive field of each point of the output map corresponds to a small patch of the input image. The size of the patch here is $64 \times 64$, which makes the discriminator pay more attention to the local details.

*3.2.3. Blur region recognition model BM*

Our blur recognition model uses the network structure of DeepLab-v3 [29]. The prefix network is ResNet50, and we take the output of the 40th activation layer of ResNet50 as the input of the subsequent ASPP module. The ASPP module is a classic structure in the DeepLab series, and thus no further description is given. The output mask of the blur recognition model, $I_b$, is stacked with $I_m$ as the input of the generator $G_\theta$, as shown in Fig. 2.

*3.3. Loss function*

The losses of our method consists of two parts: the loss of the generator $G_\theta$ and the loss of the discriminator $D_\theta$. We will introduce them separately.

$$L_{G_\theta} = L_{content} + \lambda L_{adv}, \qquad (1)$$

In formula (1), $L_G$ represents the total loss of the generator. $\lambda$ is a weight that balances the $L_{content}$ loss term and the $L_{adv}$ loss term.

The first term $L_{content}$ represents the MAE (mean absolute error) loss between the real fusion image $I_r$ and the generated fused image $I_f$. The loss is defined as follows:

$$L_{content} = \frac{1}{NCWH} \sum_{n=1}^{N} \sum_{k=1}^{C} \sum_{i=1}^{W} \sum_{j=1}^{H} |I_{r(n,k,i,j)} - I_{f(n,k,i,j)}|, \qquad (2)$$

In formula (2), $N$, $C$, $W$, and $H$ represent the batch number of generated images $I_f$ and the real image $I_r$, the number of channels, the width and the height, respectively. $I_{r(n,k,i,j)}$ represents the pixel value of the position $(i, j)$ on the kth channel of the real fused image $I_r$. $I_{f(n,k,i,j)}$ has the same meaning as $I_{r(n,k,i,j)}$, and $|.|$ is an absolute function.

The second term represents the adversarial loss of the generator $G_\theta$. We use the binary cross-entropy (BCE) to define this loss as follows:

$$L_{adv} = \frac{1}{N} \sum_{n=1}^{N} \big(- log\big( D_\theta(I_f^n)\big)\big), \qquad (3)$$

In formula (3), $N$ represents the batch number, $I_f^n$ represents the nth generated image, and $D_\theta(I_f^n)$ represents the value of the generated fused image $I_f$ that is input to the discriminator $D_\theta$.

In fact, the generator can be optimized by only $L_G$ without the discriminator. However, the reconstructed fused images can only recover the low-frequency information of the image without fine-grained textural information. Thus, we added the discriminator loss as follows:

$$L_{D_\theta} = -\frac{1}{N} \sum_{n=1}^{N} (log( D_\theta(I_r^n)) + log( 1 - D_\theta(I_f^n))), \qquad (4)$$

In formula (4), $N$ still represents the batch number, $L_G$ represents the nth real fused image, and $D_\theta(*)$ represents the output value of the discriminator. Here, we use the BCE loss $L_D$ to optimize the discriminator, thereby enabling the discriminator to well distinguish between the real fused image $I_r$ and the generated fused image $I_f$.

*3.4. Training strategy*

The complete architecture of the proposed method is shown in Fig. 2. We first train the blur recognition model BM and then train the conditional GAN model for image fusion. After the blur recognition model is well trained, its parameters are frozen and not updated in the subsequent training process. For the blur recognition model, its input is the 3-channel RGB image and its output is a 2-channel semantic segmentation mask. To reduce the workload of the manual annotation of blurred areas, this paper develops a degradation model, which can generate random local blurred areas for the input images. The artificial and real blurred images are then mixed to train the BM model. For the fusion model $G_\theta$, the input is a stacked multi-channel image of the RGB image $I_m$ and the output mask $I_b$ of the BM module. During training, we fix the parameters of the discriminator $D_\theta$ and use $L_G$ to optimize the fusion model. For the discriminator model $D_\theta$, we fix the parameters of $G_\theta$ and use $L_G$ to optimize the discriminator. In each round, we alternately train the generator $G_\theta$ and discriminator $D_\theta$ until they converge. The specific training process of the method in this paper is shown in Algorithm 1. The specific implementations of all of our networks are available at https://github.com/GengXieBo/fusion.

| *Algorithm 1:* Training procedure of Generative Adversarial Nets |
|---|
| *1* **Training BM:** |
| *2*    **for** *number of training iterations* **do** |
| *3*         *Select m blured images;* |
| *4*         *Update BM by AdamOptimizer;* |
| *5*    **end** |
| *6* **Fixed model BM parameters** |
| *7* **Training $G_\theta$ and $D_\theta$:** |
| *8*    **for** *number of training iterations* **do** |
| *9*         *Select m single-layer or few-layer images and their conresponding mask in BM and concatenate them* |

|    | $\{I_c^{(1)},...I_c^{(m)}\}$ ; |
|---|---|
| 10 | **Update** $G_\theta$ by AdamOptimzer: $\nabla_\theta L_{G_\theta}$; |
| 11 | Select m fusion patches $\{I_f^{(1)},...I_f^{(m)}\}$ from $G_\theta$; |
| 12 | Select m real fusion patchs $\{I_r^{(1)},...I_r^{(m)}\}$; |
| 13 | **Update** $D_\theta$ by AdamOptimzer: $\nabla_\theta L_{D_\theta}$; |
| 15 | End |

## 4. Experiment

To evaluate our proposed method, we will briefly introduce the evaluation metrics of fused images and compare them with traditional image fusion methods based on the transform domain and spatial domain and deep learning-based image fusion methods. The traditional methods include the wavelet transform (WT) [30], Laplacian pyramid (LPP) [31] and double-tree complex wavelet transform (DTCWT) [32]. The deep learning-based methods includes a CNN-based method for multi-focus image fusion [33] and a GAN-based method for infrared and visible image fusion (FusionGAN) [26]. In addition, we also compared our method with the classical image-to-image conditional GAN Pixel2Pixel [27] and the image super-resolution method SRGAN [34]. The above fusion methods WT, LPP and DTCWT have good MATLAB implementation codes. The implementation codes of FusionGAN, Pixel2Pixel, and SRGAN are provided by the authors. We rewrote the codes of the CNN method according to its original description. In addition, we did ablation studies to verify that the blur recognition module that we added can improve the effect of the fusion model. Finally, we applied our method to WSI slides to verify the practical application value of our proposed method. All of our experiments were run on a 12 GB GeForce RTX 2080Ti graphics card.

*4.1. Evaluation metrics*

Here, we selected three metrics to evaluate the correlation between the generated fusion images and the ground truth images: the structural similarity (SSIM) [35], the correlation coefficient (CC) [36] and the normalized mutual information ($Q_{MI}$) [37].

The SSIM is a metric to measure the similarity between two images.

$$SSIM(X,Y) = \frac{(2\mu_x\mu_y+c_1)(2\sigma_{xy}+c_2)}{(\mu_x^2+\mu_y^2+c_1)(\sigma_x^2+\sigma_y^2+c_2)}, \quad (7)$$

In formula (7), $\mu_x$ is the average value of $X$, $\sigma_x^2$ is the variance of $X$, and $\sigma_{xy}$ is the covariance of $X$ and $Y$. $c_1=(k_1L)^2$ and $c_2=(k_2L)^2$ are the constants that are used to maintain stability, where $k_1=0.01$, $k_2=0.03$. $L$ is the dynamic range of the pixel value. The SSIM ranges from 0 to 1.

The CC indicates the degree of linear correlation between two images, which is defined as follows:

$$CC(X,Y) = \frac{\sum_{i=1}^{M}\sum_{j=1}^{N}(X(i,j)-\mu_x)(Y(i,j)-\mu_y)}{\sqrt{\sum_{i=1}^{M}\sum_{j=1}^{N}(X(i,j)-\mu_x)^2(\sum_{i=1}^{M}\sum_{j=1}^{N}(Y(i,j)-\mu_y)^2)}}, \quad (8)$$

In formula (8), $M$ and $N$ are the width and height of the images $X$ and $Y$, and $\mu_x$ and $\mu_y$ are the means of $X$ and $Y$, respectively.

Mutual information is a measure of the similarity between two image intensity distributions. The mutual information is normalized to values from 0 to 1 and is defined as follows:

$$I(X,Y) = \sum_{y\in Y}\sum_{x\in X} p(x,y) \log(\frac{p(x,y)}{p(x)p(y)}), \quad (9)$$

$$H(X) = -\sum_{i=0}^{n} P(x_i) \log_2 P(x_i), \quad (10)$$

$$Q_{MI}(X,Y) = 2\frac{I(X,Y)}{H(X)+H(Y)}, \quad (11)$$

Here, $p(x, y)$ represents the joint distribution of $X$ and $Y$. $p(x)$ and $p(y)$ represent the edge distribution of $X$ and $Y$, respectively. $n$ is the maximum value of the pixels. In our experiments, $n$ is 255.

*4.2 Data and training details*

For the blur recognition model BM, there are 678 pairs of real images and ground truth masks. 600 were used the training set and 78 were used as the test set. We used the Gaussian blur degradation algorithm to generate 5000 blurred images and their corresponding masks. 4500 were used as the training set and 500 were used as the test set. During training, 600 real blurred images and 4500 degraded blurred images were mixed. For the fusion model, we used 5 cervical cell slides and scanned them under a 20× magnification optical microscope. The imaging device (3DHistech Scanner) selected a focus depth of most cells in focus as the layer 0, and then adjusted the focus depth to obtain 11 layers with the layer 0 as the middle layer. The focus depth interval is 2.7 μm and each layer image size is 86784×100352×3 (0.243 μm/pixel). We cropped 39,211 image patches with a size of 512×512×3×11 from the 5 digital slides. 37,211 were used as the training set and 2000 were used as the test set. Fig. 4 shows an example of an image patch. We used 3DHistech's CaseViewer software to obtain the fused images of the 11 layer images as the ground truth for the generator $G_\theta$.

For the blur recognition model BM, the learning rate is set to 1e-3, the batch size is set to 4, and the optimizer is Adam. We selected the parameter weight with stable effect after convergence as our blur recognition model. We trained a network that generates fused images from a single layer 0 image and a network that generates fused images from layer 0 and ±1 images. The batch size is set to 1 due to memory limitations. $\lambda$ is set to 0.001, the learning rate of generator $G_\theta$ is set to $0.5\times10^{-4}$, the learning rate of discriminator $D_\theta$ is set to half of the generator $0.25\times10^{-4}$, and the optimizer is Adam. The learning rates of the generator and discriminator are attenuated after each epoch, $lr=lr\times0.8$. Because our discriminator adopts PatchGAN, the output of the discriminator is a single channel 8×8 size feature map. Each point on the feature map corresponds to a 64×64 patch of the input image.

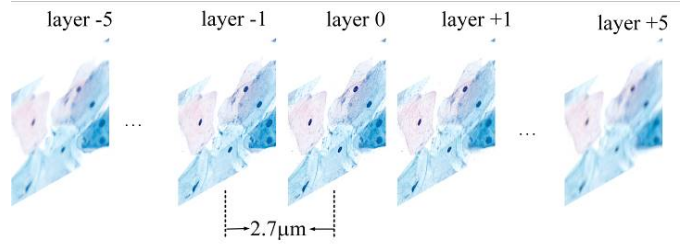

**Fig. 4**. Example images of different focus depths with a size of 512×512×3×11. 512 is the width and height, 3 is the number of channels, 11 is the number of layers, and the interval between layers is 2.7 μm.

*4.3. Comparison experiments of different methods*

*4.3.1. Test results of single layer fusion model*

In this section, we only compared our method with FusionGAN, Pixel2Pixel and SRGAN. This is because traditional image fusion methods such as WT, DTCWT and LPP require multiple images as the input. In addition, the CNN-based image fusion method also requires multiple focus images as the input of the network.

**Qualitative comparisons.** In Fig.5, the subfigures from left to right are the layer 0 image; the results of FusionGAN, Pixel2Pixel, SRAGN and our method; and the last column is the ground truth. We can observe that the input layer 0 image has partially out-of-focus nucleus and cytoplasm. The reconstruction results of the out-of-focus areas of the different models are obviously different. The generated fusion images of FusionGAN have a poor recovery effect compared to the ground truth. Pixel2Pixel and SRGAN can recover out-of-focus areas to a certain extent. However, artifacts are introduced and the image textures are not natural. Compared with the above methods, the generated fused images of our method are obviously better. Our reconstructed images contain rich textural details, and do not lose the information of the input image. Our generated fused image quality is close to the ground truth.

**Table 1**. The mean values of the evaluation metrics of the different methods on the test set.

| Metrics | FusionGAN | Pixel2Pixel | SRGAN | our |
|---|---|---|---|---|
| SSIM | 0.8414 | 0.8177 | 0.8535 | **0.8835** |
| CC | 0.9806 | 0.9807 | 0.9851 | **0.9910** |
| $Q_{MI}$ | 0.3216 | 0.3173 | 0.3386 | **0.3716** |

**Quantitative comparisons.** In Fig. 6, we quantitatively compare the four models using the three evaluation metrics on 30 randomly selected test images. The red dotted line refers to our model, and the quantified values are always greater than those of FusionGAN, Pixel2Pixel and SRGAN for the 30 test images. To ensure the generality of our quantitative test results, we further obtained the metric values for all test images (2000 images) as shown in Table 1. Our method obtained the best result for each metric. The results demonstrate that our proposed method is superior to FusionGAN, Pixel2Pixel and SRGAN.

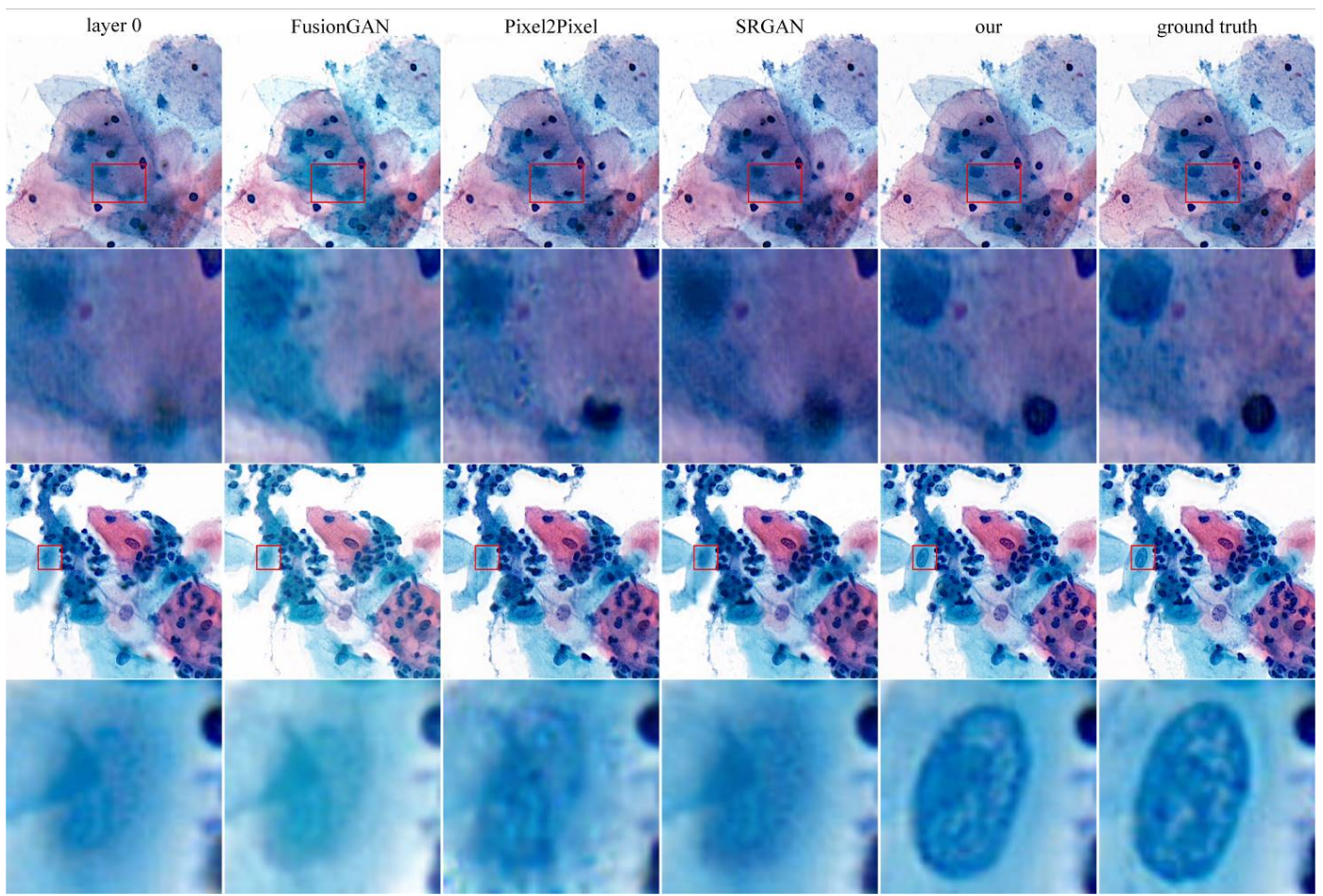

**Fig. 5.** Comparisons of the different methods' results in the single layer fusion model. The subfigures from left to right are the input layer 0 image; the outputs of FusionGAN, Pixel2Pixel, SRAGN and our method; and the last column is the ground truth. The first row and the third row in the above figure are different test images. The red rectangle is the area that is obviously blurred in the layer 0 image, and the second row and the fourth row are enlarged views of the red rectangle.

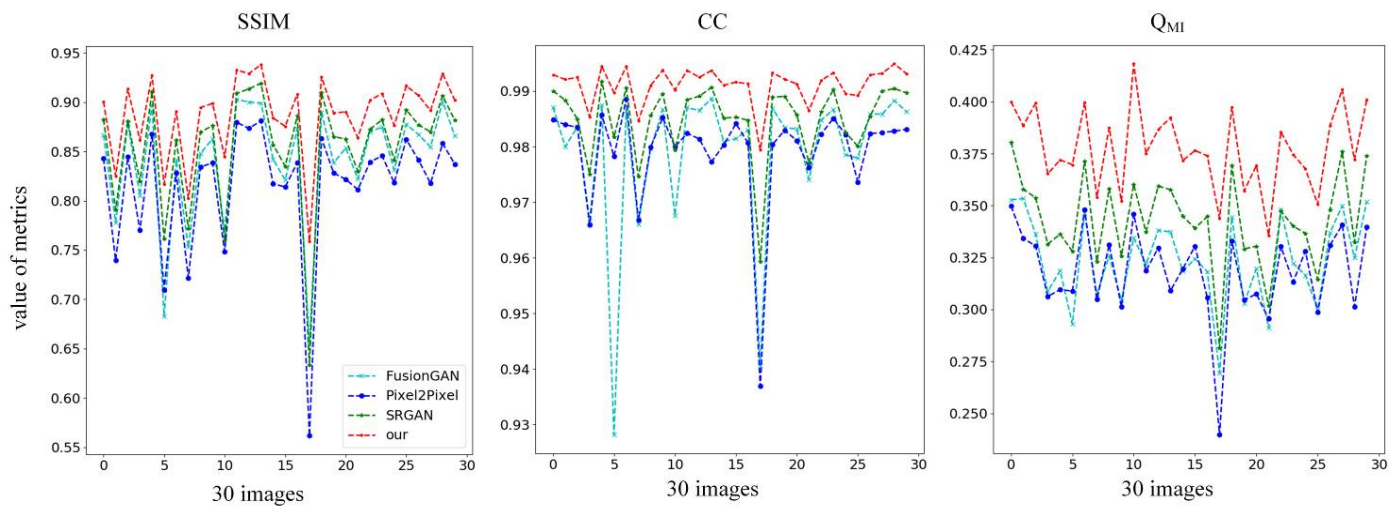

**Fig. 6.** Quantitative evaluation of the three metrics. The evaluation metrics are the *SSIM*, *CC* and $Q_{MI}$, respectively. The *x*-axis is 30 test images and the *y*-axis is the values of the metrics. The different colored polylines refer to the different methods.

*4.3.2. Test results of the few layer fusion model*

**Qualitative comparisons.** Here, the few layer fusion is compared to the single layer fusion. We compared the generated fused images from the layer 0, ±1 images of different methods. Since generating fusion images from three images of different focus depths, we added the CNN, WT, DTCWT and LPP as the compared methods. Fig. 7 shows the comparisons of the fusion images that were generated from layer 0, ±1 images by different methods. It can be observed that the different layer images have different in-focus regions. The purpose of our model is to recognize the in-focus parts of the input images, retain them in the reconstructed fused images, and to recover the out-of-focus parts as much as possible. As shown in Fig. 7, the effect of FusionGAN is worse than those of Pixel2Pixel and SRGAN. Although Pixel2Pixel and SRGAN can produce relatively clear textures, the textures are not natural and have some artifacts. The CNN-based fusion method can use the information in the input images to fuse the images as much as possible, but it does not recover the out-of-focus areas in the input images. In traditional fusion methods, DTCWT works best, but its disadvantage is the same as the CNN-based method. Although the results of our method have some differences from the ground truth, the image textures are the clearest in several methods, and the image quality is closest to the ground truth.

Table 2. The mean values of the evaluation metrics of the different methods on the test set.

| Metrics | FusionGAN | Pixel2Pixel | SRGAN | CNN | WT | LPP | DTCWT | our |
|---|---|---|---|---|---|---|---|---|
| SSIM | 0.899 | 0.885 | 0.898 | 0.894 | 0.886 | 0.864 | 0.896 | **0.938** |
| CC | 0.986 | 0.988 | 0.990 | 0.987 | 0.986 | 0.974 | 0.988 | **0.996** |
| $Q_{MI}$ | 0.356 | 0.364 | 0.368 | 0.380 | 0.349 | 0.334 | 0.364 | **0.441** |

**Qualitative comparisons.** Fig. 8 shows the three metrics values for 30 randomly selected test images with different methods. We can observe that the SSIM, CC and $Q_{MI}$ of our method are always better than those of other methods. In Table 2, we obtained the three metrics' values for the whole test set. Similarly, our method obtained the best SSIM, CC and $Q_{MI}$. Since the input 3 layer images have more information than single layer images, the metrics of the 3 layer fusion model in Table 2 are overall better than those of the single layer fusion model in Table 1. However, our method still outperforms other methods including traditional fusion methods and deep learning-based generative methods.

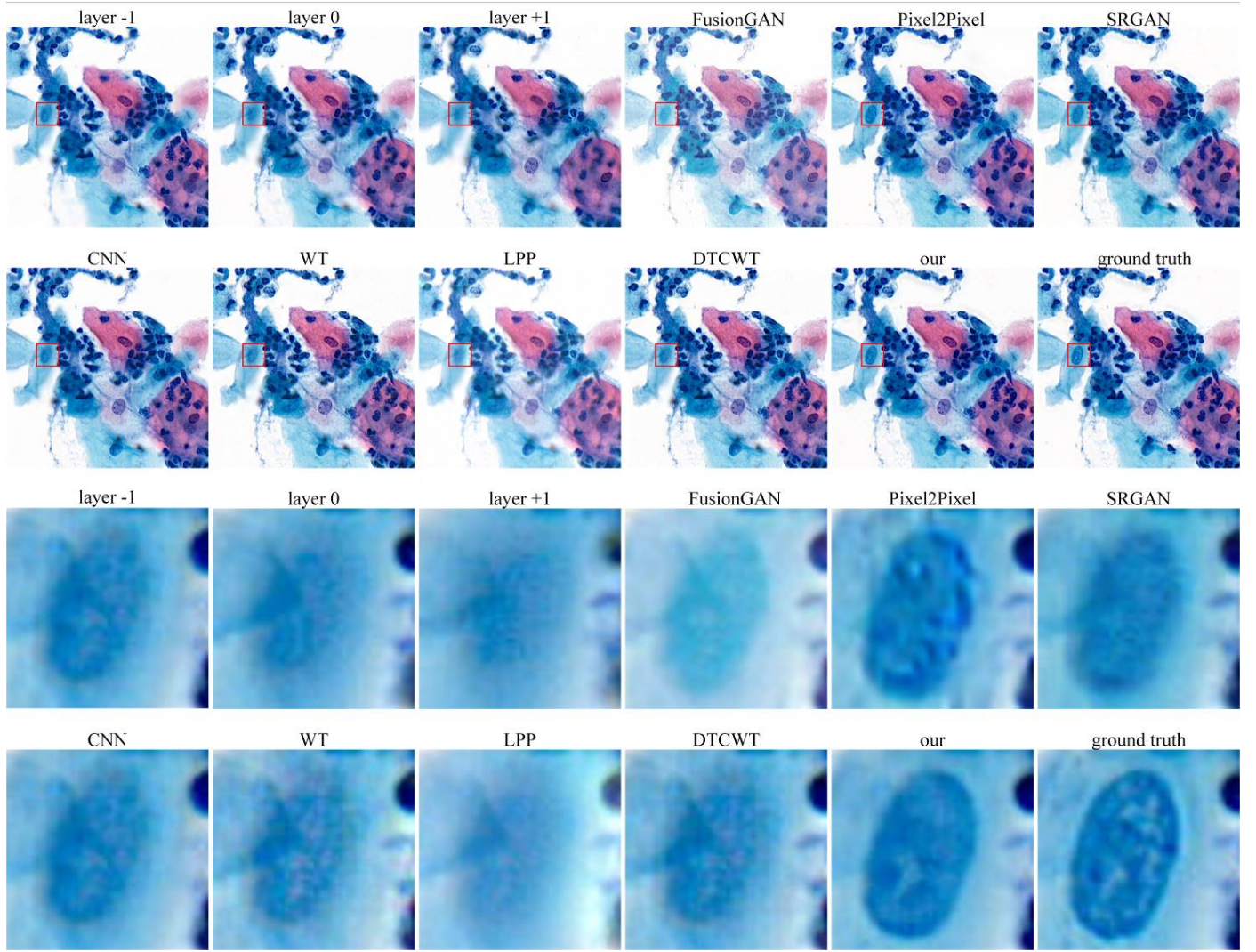

**Fig. 7**. Comparisons of the different methods' results in the three layer fusion model. The first two rows are the input layer 0 and ±1 images; the fusion images that are generated by FusionGAN, Pixel2Pixel, SRGAN, CNN, WT, LPP, DTCWT and our model; and the ground truth. The third and fourth rows are the corresponding enlarged local images.

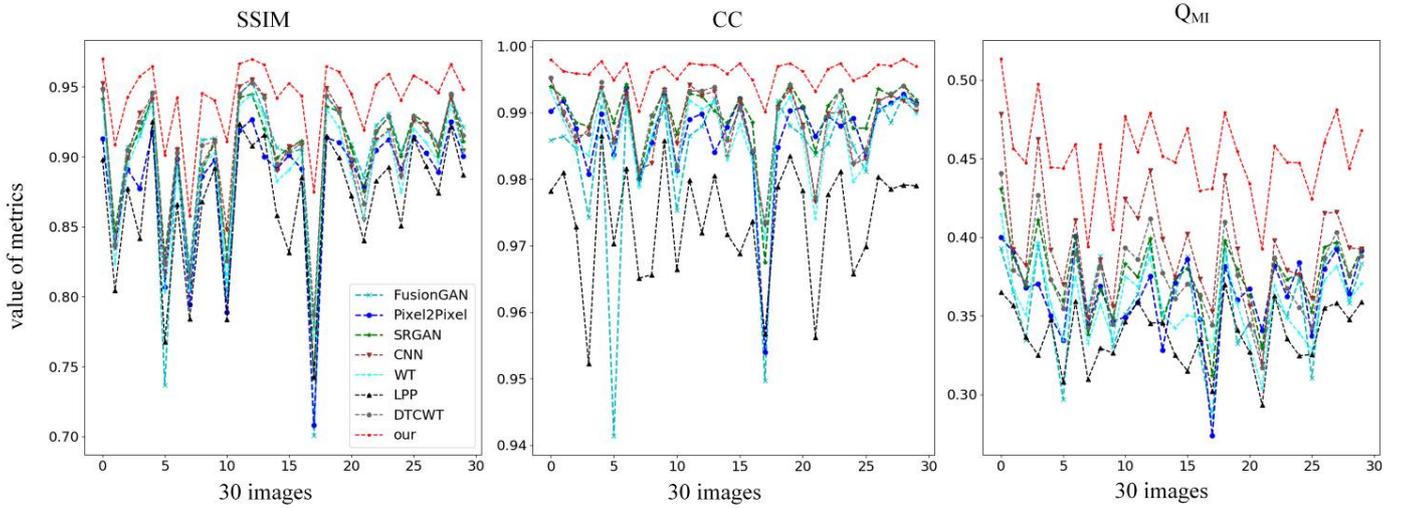

**Fig. 8**. Quantitative evaluation of the three layer fusion model for the *SSIM*, *CC* and $Q_{MI}$ metrics. The *x*-axis is 30 test images and the *y*-axis is the values of the metrics. The different colored polylines refer to the different methods.

*4.4. Ablation studies of blur recognition model*

To verify that the blur recognition model (BM) that we added is indeed valid, we performed ablation studies to verify whether the blur mask $I_b$ of the input images as an additional condition that concatenated with $I_m$ is more effective than $I_m$ alone. Here, we tested the single layer fusion model and the 3 layer fusion model using the test set. Fig. 9 shows the test results of the single layer fusion model. Both models with the BM and without the BM can generate good fusion images. However, if we look closely, we will find that the nucleuses of model with the BM are more complete and clearer than those of model without the BM. The values of the three metrics in Table 3 reflect the differences of the model with the BM and the model without the BM on the test set. Although the differences are relatively small, the BM module can help the generator to reconstruct fused images with good local textures. We believe that the blur mask can guide the generator to focus on the fusion of the blurred regions and the recovery of the fine-grained textural information.

**Table 3**. Differences between the model with the BM and the model without the BM for evaluation metrics values on the test set.

|  | Single-layer fusion | | Few-layer fusion | |
| --- | --- | --- | --- | --- |
| Metrics | our_no_BM | our_BM | our_no_BM | our_BM |
| SSIM | 0.874 | **0.885** | 0.935 | **0.938** |
| CC | 0.990 | **0.991** | 0.996 | **0.996** |
| $Q_{MI}$ | 0.362 | **0.364** | 0.432 | **0.441** |

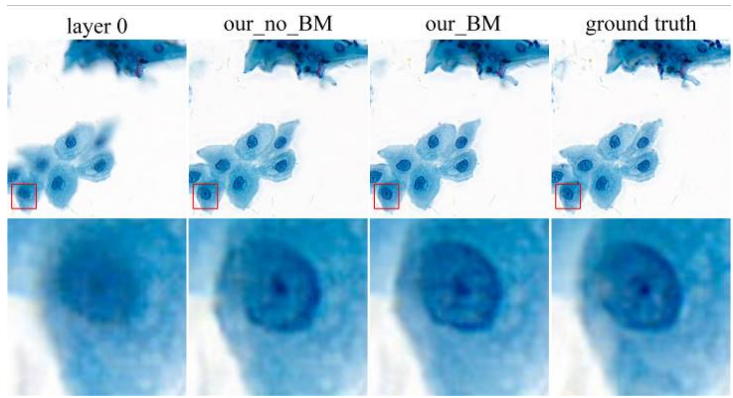

**Fig. 9**. Comparisons of the model with the BM and the model without the BM. The subfigures from left to right are the input layer 0 image, the fused image of the model without the BM, the fused image of the model with the BM and the ground truth.

*4.5. Application in WSIs*

To verify that our method can be applied to practical cytopathological WSIs, we tested a WSI $I_{ws}$ with a size of 86784×100352×3 using the single layer image fusion model. As Fig. 10 shows, we obtained a reconstructed WSI with clearer textures and a larger depth of field than the input WSI. We can observe that our model is very effective at reconstructing the out-of-focus areas. The experiment proves that our model can be applied in practice. Notably, we adopted division with redundancies when applying our method for WSIs. Specifically, the WSI $I_{ws}$ is split into many 512x512x3 image tiles with 128 pixel redundancy of neighboring tiles. After obtaining the reconstructed images of all tiles, we merged them according a distance weighting method.

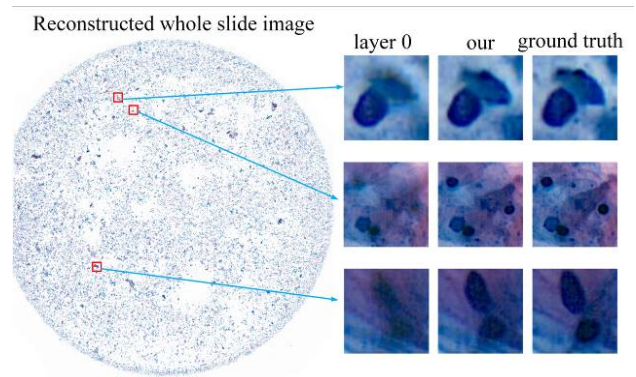

**Fig. 10**. A reconstructed WSI using single layer fusion model. The left subfigure is the reconstructed WSI by single layer fusion model. The right subfigures are the enlarged views of part blocks cropped from the reconstructed WSI.

## 5. Conclusion

In this paper, we present a method for generating a fused image from single-focus or few-focus images based on conditional GAN. This method can generate fused images with clear textures and large depth of field without requiring multiple images of different focus depths. Thus, the time and hardware costs to capture multiple focus images can be saved, which is especially important for huge sized cytopathological digital slides. Since our proposed image fusion method is an end-to-end model that can generate fused images directly from the input images, the manual design of complex activity level measurement and fusion rules is avoided. In addition, we designed a blur recognition module that can help the generator to better reconstruct fusion images with good textures. The experimental results on the test set show that our method outperforms the traditional image fusion algorithms, recent GAN-based image fusion methods and classical image-to-image condition GAN methods.

The proposed method can improve the quality of the large-size digital cytopathological slides when time and hardware costs are limited. This is of great help to medical staff to further interpret digital slides or conduct automated algorithmic analysis. For clinical applications, our model has some disadvantages. Our model is trained and tested on a single style of dataset. However, the styles of digital slides are diverse due to the different stains and imaging equipment. The quality of the fused images that are generated by our model on other styles of slides may decline. Scanning various styles of slides and refining the model will cost a lot. In the future, we will address this challenge by generalizing our model to different datasets in an unsupervised way.

**Acknowledgment**

We thank the Optical Bioimaging Core Facility of WNLO HUST for the support in data acquisition.